%% file: acl_latex.tex
\pdfoutput=1

\documentclass[11pt]{article}

\usepackage[]{acl}

\usepackage{times}
\usepackage{latexsym}

\usepackage[T1]{fontenc}

\usepackage[utf8]{inputenc}

\usepackage{microtype}

\usepackage{inconsolata}

\usepackage{graphicx}

\usepackage{booktabs}

\usepackage{subcaption}

\usepackage{float}
\usepackage{multirow}
\usepackage{tcolorbox}
\usepackage{listings}
\usepackage{xcolor}
\usepackage{caption}

\lstdefinestyle{pythonstyle}{
  language=Python,
  basicstyle=\ttfamily\small,
  keywordstyle=\color{blue},
  commentstyle=\color{gray},
  stringstyle=\color{purple},
  showstringspaces=false,
  tabsize=4,
  breaklines=true,
  frame=single,
  framerule=0.5pt,
  backgroundcolor=\color{gray!5},
  rulecolor=\color{gray!50},
  xleftmargin=0.5em,
  xrightmargin=0.5em
}

\usepackage{stfloats}
\usepackage{listings}
\usepackage{xcolor}  

\title{From Curated Data to Scalable Models: Continual Pre-training of Dense and MoE Large Language Models for Tibetan}

\author{
    Lei Yang\footnotemark[1],
    Leiyu Pan\footnotemark[1],
    Bojian Xiong,
    Renren Jin,
    Shaowei Zhang \\
    \textbf{Yue Chen},
    \textbf{Ling Shi},
    \textbf{Jiang Zhou},
    \textbf{Junru Wu},
    \textbf{Zhen Wang},
    \textbf{Jianxiang Peng} \\
    \textbf{Juesi Xiao},
    \textbf{Tianyu Dong},
    \textbf{Zhuowen Han},
    \textbf{Zhuo Chen},
    \textbf{Yuqi Ren},
    \textbf{Deyi Xiong}\footnotemark[2]\\
    TJUNLP Lab, School of Computer Science and Technology, Tianjin University, China \\
    \texttt{\{yanglei\_9, lypan, dyxiong\}@tju.edu.cn}
}
\begin{document}

\maketitle

\renewcommand{\thefootnote}{\fnsymbol{footnote}} 

\footnotetext[1]{These authors contributed equally to this work.}
\footnotetext[2]{Corresponding author.}

\renewcommand{\thefootnote}{\arabic{footnote}} 

\begin{abstract}

\input{abstract}

\end{abstract}

\input{section1}

\input{section2}

\input{section3}

\input{section4}

\input{section5}

\input{section6}

\input{section7-1}

\input{section7-2}

\input{section8}

\input{section9}

\bibliography{custom}

\clearpage

\appendix

\input{appendix}

\end{document}

%% file: abstract.tex
Large language models (LLMs) have achieved remarkable success across a wide range of natural language processing tasks, yet their performance remains heavily biased toward high-resource languages. Tibetan, despite its cultural significance and large speaker population, is still substantially underrepresented. In this work, we present a comprehensive pipeline for advancing Tibetan language modeling through large-scale data curation and continual pre-training.
%
%
We construct a 72 GB high-quality Tibetan corpus, the largest to date, and adapt Qwen2.5-7B through balanced multilingual continual pre-training with Tibetan, Chinese, and English, followed by multilingual instruction tuning. To further scale capacity efficiently, we extend the dense model to a 50B-A10B Mixture-of-Experts architecture. Due to the absence of standardized Tibetan benchmarks, we build multiple evaluation datasets via high-quality translation and human verification. Experimental results show that both dense and MoE models consistently outperform existing open-source and Tibetan-focused models of similar scale across diverse tasks. Our work advances Tibetan-centric LLM research and provides transferable insights for extending LLMs to other low-resource languages.
We will release the model weights, evaluation benchmarks, and detailed data processing documentation in the follow-up.

%% file: section1.tex
\section{Introduction}


In recent years, large language models (LLMs) have achieved remarkable progress, demonstrating exceptional performance~\cite{DBLP:journals/corr/abs-2309-15025, DBLP:journals/corr/abs-2310-19736} across a wide range of natural language processing tasks~\cite{grattafiori2024llama, yang2025qwen3, DBLP:journals/corr/abs-2601-21476}, largely due to access to massive, high-quality training data~\cite{kaplan2020scaling, wettig2024qurating}. However, the pre-training corpora of mainstream LLMs are predominantly composed of high-resource languages such as English and Chinese~\cite{touvron2023llama, nie2026attnpo, yu2026knowrl}. This imbalance in data distribution has led to a strong performance bias toward high-resource languages, while support for low-resource languages remains significantly limited~\cite{zhu2024multilingual, sun2024fuxitranyu}.

The root causes of this disparity lie in the inherent scarcity of accessible data for low-resource languages, as well as the considerable challenges involved in data collection, cleaning, and standardization~\cite{kudugunta2023madlad, penedo2024fineweb}. These limitations hinder the development and effectiveness of LLMs for underrepresented languages, thereby preventing speakers of those languages from benefiting equally from the latest advances in artificial intelligence~\cite{nguyen2024seallms, dou2024sailor}. Moreover, this technological imbalance risks further concentrating linguistic resources and innovation around dominant languages, posing a serious threat to global linguistic diversity~\cite{ustun2024aya} and cultural preservation~\cite{xu2024self, li2024culturellm}. Therefore, advancing research on LLMs for low-resource languages is not only a matter of technological development but also a critical step toward promoting artificial intelligence equity and safeguarding cultural diversity~\cite{shen2024language}.


Tibetan, as a representative low-resource language, is one of China’s minority languages and is spoken by millions of native speakers. It serves as a vital vehicle for the preservation of Tibetan culture and historical heritage~\cite{liang2024tibetan, huang2025sun}. Despite its significant population base and cultural value, Tibetan remains severely underrepresented in terms of digitized linguistic resources and has long been categorized as a low-resource language~\cite{liu2022tibert, liu2025survey}. In recent years, the development of multilingual LLMs has led to gradual improvements in support for Tibetan~\cite{liu2024deepseek, lv2025t}. However, due to the dominance of high-resource languages in training corpora and the limited scale, quality, and standardization of Tibetan data, existing models still struggle with tasks involving Tibetan~\cite{wenhao2024tilamb}.

\begin{figure*}[t]
    \centering
    \includegraphics[width=0.9\textwidth]{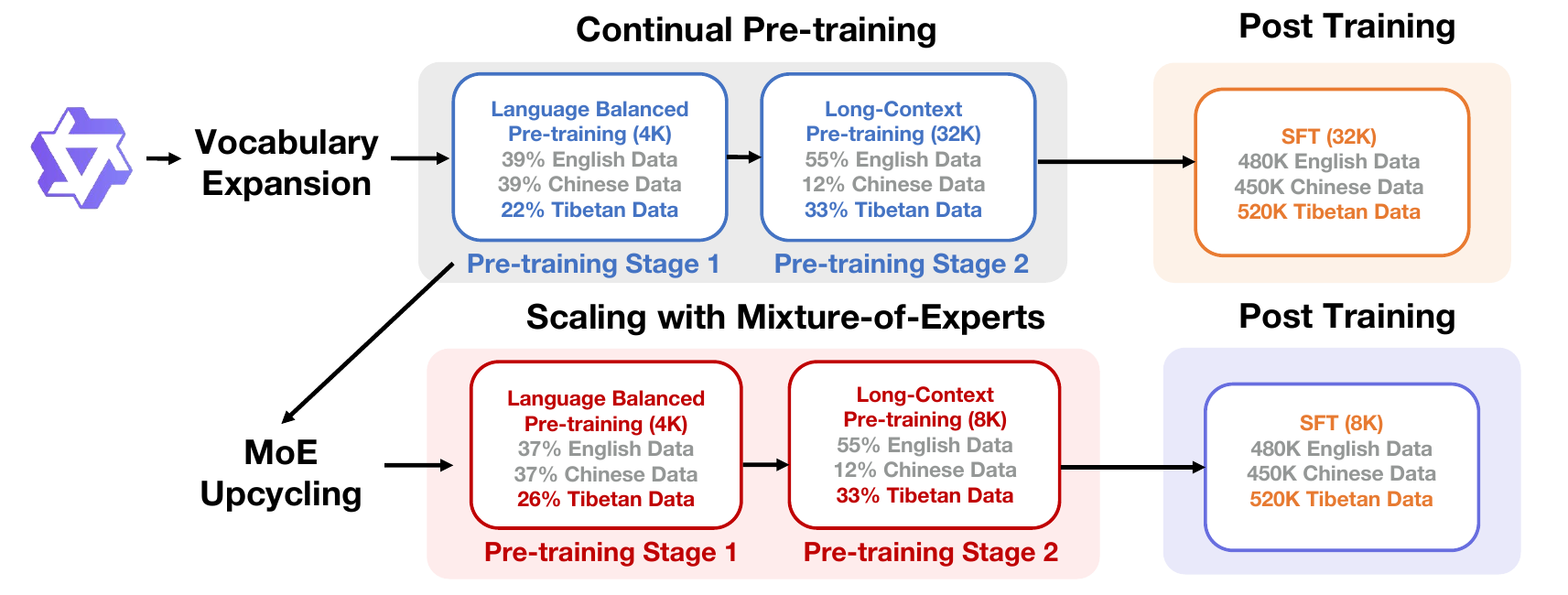}
    \caption{Depiction of the training flow for our model.}
    \label{fig:pipeline}
\end{figure*}


In view of the above challenges, we create a multilingual LLM supporting Tibetan, Chinese, and English. We present a comprehensive pipeline for building a Tibetan-capable LLM, covering data curation, pre-training, post-training, and evaluation.

To curate pre-training data for Tibetan, we build our Tibetan pre-training corpus from four primary sources: (1) publicly available datasets, (2) large-scale web crawling, (3) synthetic data generation, (4) proprietary data. To ensure high data quality, we also develop a customized data cleaning pipeline tailored for Tibetan. In total, we have collected 72 GB of cleaned Tibetan text, which, to the best of our knowledge, constitutes the largest Tibetan corpus ever constructed for LLM pre-training.


To build our model, we continue to pre-train Qwen2.5-7B-base model~\cite{DBLP:journals/corr/abs-2412-15115}, a strong open-source multilingual LLM, to extend its capabilities to Tibetan. Figure~\ref{fig:pipeline} shows our data composition and training process. To facilitate effective cross-lingual transfer and accelerate the model's adaptation, we incorporate both Chinese and English data alongside the curated Tibetan data during this continual pre-training phase. Following pre-training, we perform instruction tuning using multilingual instruction-following datasets collected from open-source repositories in Tibetan, Chinese and English, further enhancing ability of the model to handle diverse downstream tasks across the three languages.

Beyond dense model adaptation, we further scale the model using a Mixture-of-Experts (MoE) architecture~\cite{DBLP:conf/iclr/ShazeerMMDLHD17, DBLP:journals/jmlr/FedusZS22}, enabling a substantial increase in model capacity while maintaining efficient inference. This MoE-based scaling strategy allows the model to better leverage additional Tibetan data and significantly improves performance after instruction tuning.

For evaluation, given the limited availability of standardized Tibetan benchmarks for large language models, we construct multiple new Tibetan evaluation datasets by automatically  translating existing benchmarks into Tibetan. All translations are rigorously proofread by native Tibetan speakers. The translated benchmarks are supplemented with existing public Tibetan benchmarks to provide a more comprehensive evaluation framework. Results show that both our base and chat models consistently outperform other models of similar scale across a variety of tasks, and in some cases, achieve performance comparable to significantly larger LLMs. Additionally, by extending the model vocabulary to include frequent Tibetan character combinations, we further enhance both generation quality and decoding efficiency, enabling faster inference under identical runtime conditions.


Our main contributions can be summarized as follows:

\begin{itemize}
    \item We enhance the foundation for Tibetan natural language processing by constructing a high-quality Tibetan pre-training corpus (the largest Tibetan dataset to date) and developing several evaluation benchmarks specifically designed for Tibetan LLMs.
    \item With the curated data, we build 7B dense and 50B-A10B MoE multilingual LLMs with strong capabilities in Tibetan.
    \item Extensive evaluations show that our model achieves strong improvements in accuracy compared to other open-source models and Tibetan-tailored models of similar scale across various benchmarks.
\end{itemize}

%% file: section2.tex
\section{Related Work}


In recent years, multilingual LLMs have made significant progress, gradually evolving from supporting a limited number of high-resource languages to covering a broader range of low- and medium-resource languages~\cite{yang2025qwen3}. This improvement is largely driven by the continued expansion of multilingual corpora, both in terms of the number of supported languages and the amount of data available for each language~\cite{penedo2024fineweb-2, zhou2026earlvr}. As a result, the multilingual capabilities of LLMs have seen steady improvements~\cite{ustun2024aya, ji2024emma}. However, despite this progress, current multilingual models still provide insufficient support for Tibetan~\cite{DBLP:journals/corr/abs-2412-15115}. Existing open-source multilingual corpora include very limited Tibetan data in terms of both scale and diversity, which significantly hampers model performance on Tibetan language understanding and generation~\cite{kudugunta2023madlad, nguyen2024culturax}. To address this gap, we focus on building a large-scale, high-quality Tibetan pre-training corpus and training a dedicated language model with enhanced support for Tibetan, thus improving performance in downstream Tibetan tasks.

To efficiently leverage language-specific data for building models tailored to a particular language, a straightforward approach is to combine the collected data with multilingual corpora and train a new model from scratch~\cite{ji2024emma}. However, this method is computationally expensive. As a result, prior research has proposed several low-cost alternatives to enhance existing models' performance on specific languages. One line of work explores in-context learning techniques~\cite{cahyawijaya2024llms}. While these methods are relatively efficient, they do not fundamentally improve the model’s inherent capabilities in the target language and typically assume that the base model already has a reasonable degree of support for that language. This limitation makes such methods less effective for truly low-resource languages like Tibetan, where existing models perform poorly to begin with. Another line of work adopts continued pre-training to adapt pre-trained models to underrepresented languages. Notable examples include Sailor~\cite{dou2024sailor} and LLaMaTurk~\cite{toraman2024llamaturk}, which demonstrate that extending multilingual models with targeted continual pre-training can effectively enhance their capabilities in specific languages. Inspired by this approach, we employ a continued pre-training approach to enhance the Tibetan language support of the Qwen2.5 model.

Several large language models with support for Tibetan understanding and generation have been proposed recently, e.g., TiLamb~\cite{wenhao2024tilamb}, Sunshine~\cite{huang2025sun}, and T-LLaMa~\cite{lv2025t}. Some of these models adopt parameter-efficient fine-tuning techniques like LoRA for continued pre-training~\cite{DBLP:conf/iclr/HuSWALWWC22}. In contrast, our approach uses full-parameter fine-tuning, offering greater flexibility and potential for adaptation. Moreover, compared to the aforementioned models, our Tibetan training corpus is larger in scale, richer in source diversity, and covers a wider range of linguistic scenarios. As a result, the model we trained on this corpus achieves superior performance on both our curated Tibetan benchmarks and existing public evaluation datasets.

%% file: section3.tex
\begin{table}[t]
\centering
\resizebox{\columnwidth}{!}{%
\begin{tabular}{lll}
\toprule
\multicolumn{1}{l}{\textbf{Data Source}} & \multicolumn{1}{l}{\textbf{Language}} & \multicolumn{1}{l}{\textbf{Size (After Filtering)}} \\ \midrule
Open-source data                         & Tibetan                               & 9.53 G                                           \\
Web data crawling                        & Tibetan                               & 4.18 G                                           \\
Synthetic data                           & Tibetan, Chinese, English             & 17.46 G                                          \\
Private data                             & Tibetan                               & 41.13 G                                          \\ \bottomrule                                       
\end{tabular}
}
\caption{Overview of the four main sources of Tibetan training data.}
\label{tab:tibetan data composition}
\end{table}

\section{Pre-training Data Curation}


During the pre-training data curation phase, we collect Tibetan data from a wide range of sources. In addition to gathering all available open-source Tibetan corpora, we also include web-scraped content, synthetically generated data, and a portion of proprietary resources. To ensure high data quality, we design and apply a dedicated data cleaning pipeline specifically for Tibetan, which is used to process the entire dataset.

\subsection{Data Collection}



We collect a diverse and high-quality training corpus from four major sources: open-source datasets, web data crawling, synthetic data generation, and private data. These data sources vary in style, domain coverage, and linguistic features, and together form a comprehensive Tibetan pre-training corpus. Table~\ref{tab:tibetan data composition} summarizes the detail information of each data component.

\paragraph{Open-Source Data Collection} We extensively collect Tibetan data from various open-source multilingual datasets, as well as from monolingual Tibetan dataset. Detailed information about the collected datasets is provided in Appendix~\ref{appendix:open-source-data-collection-details}. The majority of the data originates from web sources, but it also includes content from other domains such as news, books, and Wikipedia, contributing to the overall diversity of the corpus. 

\paragraph{Web Data Crawling}


Although the open-source data we have collected includes a wealth of web-based content, it suffers from limited timeliness, with many recent materials not reflected in existing datasets. To address this limitation, we utilize the crawl4ai tool\footnote{\href{https://github.com/unclecode/crawl4ai}{https://github.com/unclecode/crawl4ai}} to collect data from 55 Tibetan-language websites, resulting in a total of 204,125 webpages. The crawled data spans a wide range of domains, which are primarily on news, but also include content from books, Wikipedia, and other categories, contributing to the diversity of our Tibetan corpus. Detailed information of web data crawling is given in Appendix~\ref{appendix:web-data-crawling-details}.

\paragraph{Synthetic Data Generation}


Synthetic data is often used to compensate for the lack of real-world data in low-resource settings, and it has been shown to effectively improve the performance of LLMs~\cite{abdin2024phi}. Since Tibetan is a low-resource language with limited high-quality data, we employ automatic data synthesis to supplement our Tibetan corpus. Given that existed LLMs exhibit limited ability in Tibetan monolingual generation, we choose to generate Tibetan data via translation. To further promote cross-lingual transfer, we construct parallel corpora using both source language sentences and their Tibetan translations for training. We sample a portion of the FineWeb-Edu~\cite{penedo2024fineweb} and Cosmopedia~\cite{benallal2024cosmopedia} pre-training datasets, translate them into Tibetan using the Google Translate API, and organize the parallel data using the instruction templates introduced by~\citet{zhu2024preference}.

\paragraph{Private Data Usage}


In addition to publicly available data, we also utilize a set of private Tibetan-language digital resources, primarily consisting of books and classical materials. These resources are obtained from non-public but legally accessible sources, such as internal digital archives and institutional collaborations. All data used in this category contain no personally identifiable information or user-generated content, and are solely used for research and model training purposes.

\begin{table}[t]
\centering
\resizebox{\columnwidth}{!}{%
\begin{tabular}{llll}
\toprule
\textbf{Language}                 & \textbf{Category}             & \textbf{Volume} & \textbf{Token} \\ \midrule
\multirow{7}{*}{English} & Books                & 9.8 G   & 2.77 B\\
                         & Code                 & 9.7 G   & 4.42 B \\
                         & Encyclopedia         & 3 G     & 0.77 B \\
                         & Mathematical Corpora & 4.6 G   & 1.27 B \\
                         & Academic Papers      & 5.8 G   & 1.47 B \\
                         & QA Forums            & 2.7 G   & 0.90 B \\
                         & Web Pages            & 13.9 G  & 3.35 B \\ \midrule
\multirow{4}{*}{Chinese} & Books                & 8.5 G   & 2.72 B \\
                         & Encyclopedia         & 3.3 G   & 1.00 B \\
                         & QA Forums            & 3 G     & 0.88 B \\
                         & Web Pages            & 35.1 G  & 10.23 B \\ \midrule
\multirow{2}{*}{Tibetan} & Books                & 7.5 G   & 0.88 B \\ 
                         & Web Pages            & 64.8 G  & 7.61 B \\
\bottomrule
\end{tabular}
}
\caption{Pre-training data volume for three languages.}
\label{tab:data-composition}
\end{table}

\subsection{Data Cleaning and Deduplication}


After collecting the Tibetan corpus, we perform a comprehensive data cleaning process to ensure high data quality suitable for LLM training. Due to the unique characteristics of Tibetan script, we develop a customized data cleaning pipeline tailored to Tibetan, inspired by the FineWeb data processing workflow~\cite{penedo2024fineweb}. Our cleaning pipeline consists of two main stages: language filtering and quality filtering. For language filtering, we use the FastText language classifier~\cite{grave2018learning} to identify Tibetan text, removing any documents with a Tibetan confidence score below 0.5. For quality filtering, we apply a combination of five filtering strategies: Gopher duplicate filtering, Gopher quality filtering~\cite{rae2021scaling}, C4 quality filtering~\cite{raffel2020exploring}, FineWeb quality filtering and sensitive content filtering. Some of the heuristic rules in these procedures are slightly modified to better accommodate the characteristics of Tibetan text. Detailed descriptions of quality filtering are provided in Appendix~\ref{appendix:quality-filter-details}.

After completing the data cleaning process, we apply MinHash-based deduplication using the DataTrove tool\footnote{\href{https://github.com/huggingface/datatrove}{https://github.com/huggingface/datatrove}} to remove highly redundant text segments. This step helps enhance the diversity and effectiveness of the pretraining data.

\subsection{Data Composition}


After cleaning and deduplicating the Tibetan data, we further enhance the model’s Tibetan capabilities by incorporating high-quality, multi-domain Chinese and English data~\cite{chen2024towards} into the continued pre-training process. This approach leverages cross-lingual generalization to transfer the model’s existing strengths in Chinese and English to Tibetan. Additionally, it functions as a data replay strategy, helping to prevent catastrophic forgetting of general knowledge that could arise if the model were trained exclusively on Tibetan data.
Additionally, we adopt a balanced multilingual data ratio, aiming to maximize the model’s Tibetan performance by mitigating interference from high-resource languages while still benefiting from cross-lingual generalization. The data volumes used for continued pre-training in each language are summarized in Table~\ref{tab:data-composition}.

%% file: section4.tex
\begin{figure}[t]
    \centering
    \includegraphics[width=\columnwidth]{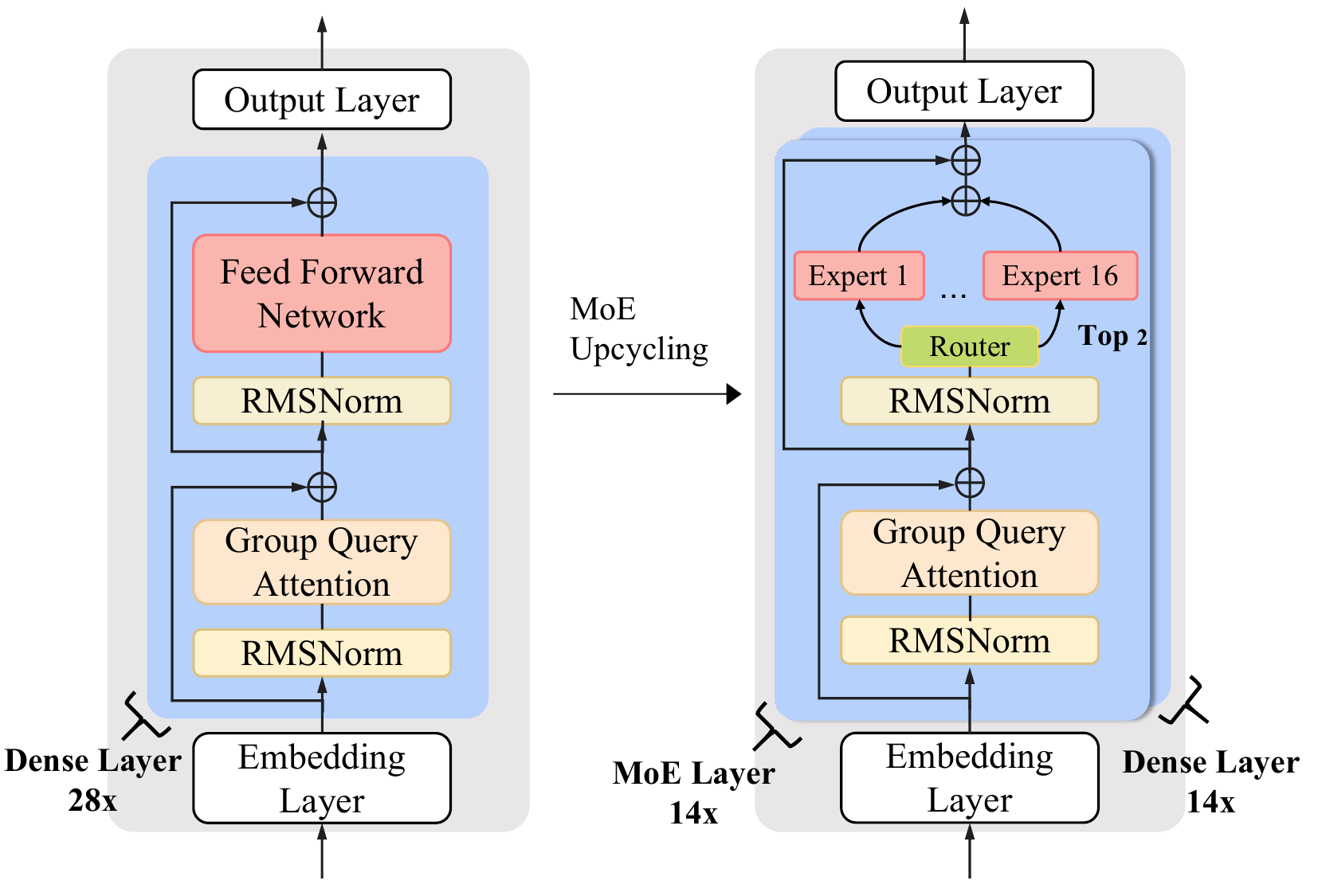}
    \caption{Architecture of the dense and MoE models. In the MoE model, each expert is initialized from the Feed Forward Network parameters of the dense model, while the router parameters are randomly initialized.}
    \label{fig:model}
\end{figure}

\section{Continual Pre-training}


We select Qwen2.5-7B-base~\cite{DBLP:journals/corr/abs-2412-15115} as our foundation model due to its strong multilingual capabilities, which provides a solid foundation for cross-lingual transfer to low-resource languages such as Tibetan. To enhance its ability to process Tibetan text, we first expand the vocabulary to include commonly used Tibetan character combinations, enabling more efficient and accurate tokenization. We then adopt a two-stage continued pre-training strategy: a language-balanced pre-training stage with a short context window to transfer Chinese and English knowledge to Tibetan, followed by a long-context pre-training stage to extend the model's context length. We show the composition of the data and the overall training process in Figure~\ref{fig:pipeline}, and present the overall architecture of the model in Figure~\ref{fig:model}.

\subsection{Vocabulary Expansion}


The original Qwen2.5 tokenizer tends to over-segment Tibetan characters, resulting in higher computational costs and reduced generation controllability. To address this, we extend its vocabulary by training a 15K Byte-level BPE tokenizer on a representative Tibetan corpus and merging it with the original vocabulary, removing duplicates. The final vocabulary contains 165,428 tokens, enabling more efficient and accurate Tibetan tokenization. Furthermore, we conduct a detailed comparison of the tokenization efficiency of various models in Section~\ref{sec:token}, and the results show that our tokenizer has significant advantages.

\subsection{Stage 1: Language Balanced Pre-training}
\label{sec:stage1}


The first stage of continued pre-training aims to build foundational Tibetan capabilities by leveraging both Tibetan monolingual data and cross-lingual transfer from high-resource languages. To ensure knowledge retention and avoid catastrophic forgetting, we use a balanced mix of Chinese, English, and Tibetan data during training.

After tokenization using our extended tokenizer, the training dataset comprises approximately 15B Chinese tokens, 15B English tokens, and 8.5B Tibetan tokens. It is worth noting that although the raw volume of Tibetan data is substantial, the final token count is lower due to the higher compression efficiency of our tokenizer, which is specifically adapted for Tibetan.

In this stage, we perform continual pre-training starting from the Qwen2.5-7B-base model, preserving its original configuration. To improve training efficiency, we use a relatively short context window of 4K tokens. Additional training details are provided in Appendix~\ref{continual-pre-training-details}.

\subsection{Stage 2: Long Context Pre-training}
\label{sec:stage2}


To enhance the model's capability for processing long inputs~\cite{DBLP:conf/emnlp/YangXPZX25}, we conduct long context pre-training based on the first stage model to extend the context window to 32K tokens.


The training data consist of trilingual mixed text in Chinese (0.56B), English (2.5B), and Tibetan (1.5B, which is not utilized in the first stage), totaling 4.56B tokens. This contributes to improving the model's generalization capability for Tibetan long contexts. Regarding training parameters, we maintain consistency with Qwen2.5-7B-base and the first stage pre-training. Specific data distribution and training parameters are detailed in Section~\ref{long_context_pre-training_details}.


We evaluate the model's performance in long context comprehension through needle-in-a-haystack evaluation using Claude as the evaluator. Due to the scarcity of Tibetan data, we construct a needle-in-a-haystack dataset for Tibetan text and employ recent news articles as needles to avoid the influence of the model's inherent knowledge on the evaluation. Experimental results demonstrate that the model after long context pre-training significantly outperforms the model that completed only the first stage pre-training. Detailed experiments are provided in Section~\ref{long_context_pre-training_details}.

%% file: section5.tex
\section{Post Training}
\label{sec:post_training}

After continual pre-training, the model has acquired foundational capabilities in Tibetan. To further enhance its ability to follow instructions, we perform instruction tuning using a diverse and richly curated set of instruction data. In terms of domain coverage, the open-source instruction data we have collected spans a wide range of areas, including instruction following, multi-turn dialogue, translation, summarization, etc. This diverse coverage is designed to comprehensively enhance the model’s alignment capabilities across different tasks. From a linguistic perspective, to facilitate the transfer of instruction-following capabilities from Chinese and English to Tibetan, we construct a multilingual instruction dataset that covers three languages. Detailed information on the post-training data and process is provided in Appendix~\ref{post_training_details}.




%% file: section6.tex
\section{Scaling with Mixture-of-Experts}


To further expand model capacity while maintaining efficient inference and to explore scalable modeling pathways for low-resource language scenarios, we investigated a Mixture-of-Experts (MoE) architecture specifically tailored for Tibetan language modeling. Our training data, processes, and architecture are shown in Figure~\ref{fig:pipeline} and Figure~\ref{fig:model}.

\subsection{MoE Architecture and Data}


We train the MoE model through an upcycling approach starting from a 7B-scale dense model~\cite{DBLP:conf/nips/ZhangGG0CVFBRUL24, DBLP:journals/corr/abs-2410-07524, DBLP:conf/acl/HuiZW00S25, DBLP:conf/icml/LiewKT25}. Specifically, we set the MoE layer frequency to 2, meaning that every other layer replaces the original feed-forward network with an MoE layer, while the self-attention structure remains unchanged. Each MoE layer comprises 16 independent experts, and during inference, a Top-2 gating mechanism is employed, whereby each token activates 2 experts.


Under this configuration, the MoE model contains a total of approximately 50B parameters, while the number of parameters actually activated per token during inference is approximately 10B. This design significantly enhances the model's expressive capacity while maintaining high inference efficiency~\cite{DBLP:journals/corr/abs-2503-07137, DBLP:conf/naacl/LoHQWF25}.


Beyond architectural scaling, MoE training is accompanied by data expansion. Building upon the 7B dense model training corpus, we additionally incorporate approximately 2.2B tokens of data, primarily consisting of Tibetan PDF documents and Chinese texts related to Tibetan content. These data predominantly feature long-form, high-quality texts that align with the MoE model's larger capacity and contextual modeling capabilities, thereby facilitating improvements in knowledge coverage and long-text comprehension.

\subsection{Training Pipeline}


This section focuses on describing the initialization strategy and training pipeline of the MoE model. Our primary MoE model is not initialized directly from a general-purpose multilingual foundation model, but rather from the dense Tibetan-adapted checkpoint described in Section~\ref{sec:stage1}. This design choice ensures that MoE experts inherit robust and well-aligned Tibetan language representations before undergoing further specialization.


Regarding the training pipeline, we first conduct continued pre-training of the MoE model under a 4K context window, training on approximately 40.7B tokens. Subsequently, following the configuration in Section~\ref{sec:stage2}, we extend the model's context length to 8K to enhance its long-text modeling capability. Finally, we perform supervised SFT on the MoE model using the same multilingual instruction dataset as the Section~\ref{sec:post_training}, yielding the final MoE-SFT model.

\subsection{Initialization Strategy}


It is worth noting that we also explore an alternative initialization approach, which is to directly initializing the same MoE architecture from the Qwen2.5-7B-base checkpoint and conducting continued pre-training under identical settings. However, this approach fail to yield stable performance improvements, with overall performance significantly inferior to that of the MoE model initialized from the Tibetan-adapted dense model.


This observation suggests that the initialization strategy plays a critical role in MoE scaling for low-resource languages. Directly applying MoE to general-purpose multilingual foundation models may introduce fragmented language representations across different experts, thereby requiring substantially more training data to achieve effective expert specialization. In contrast, starting from a dense model that has already undergone language adaptation allows for more comprehensive inheritance of existing linguistic structures and knowledge, providing a more stable and efficient foundation for subsequent MoE scaling. We defer further analysis of expert behavior and routing dynamics under this configuration to future work.













%% file: section7-1.tex
\begin{table*}[ht]
\centering
\resizebox{\textwidth}{!}{%
\begin{tabular}{lccccccc}
\toprule
\multirow{2}{*}{\textbf{Model}} & \multicolumn{2}{c}{\textbf{Hellaswag-bo}} & \textbf{ARC-bo} & \textbf{Xcope-bo} & \textbf{Xstorycloze-bo} & \multicolumn{2}{c}{\textbf{TibetanQA}} \\
 & (acc) & (acc\_norm) & (acc) & (acc) & (acc) & (EM) & (F1) \\
\midrule
Qwen3-8B                        & 27.07  & 30.56  & 29.16  & 50.40  & 50.37  & 17.80  & 30.22  \\
LLaMA3.1-8B-Instruct            & 27.28  & 32.00  & 29.31  & 51.60  & 51.69  & 36.09  & 53.04  \\
Qwen2.5-7B-base                 & 26.69  & 31.57  & 29.60  & 51.80  & 49.97  & 19.31  & 32.38  \\
Qwen2.5-7B-Instruct             & 26.46  & 31.03  & 28.47  & 51.60  & 50.43  & 10.15  & 18.42  \\
DeepSeek-R1-Distill-Llama-8B    & 25.96  & 30.67  & 30.38  & 50.40  & 48.50  & 9.88  & 17.99  \\
\midrule
Ours-Base           & 30.30  & 36.20  & 44.32  & \textbf{59.80}  & 60.80  & \textbf{49.43} & \textbf{66.15} \\
Ours-Base-32k                & 30.09  & 36.58  & 44.62  & 58.60  & 60.60  & 38.42  & 55.52   \\
Ours-SFT       & \textbf{30.64} & \textbf{36.71} & \textbf{48.39} & 57.80  & \textbf{61.86} & 46.15  & 63.15  \\
\midrule
Ours-MoE-Base & 29.54 & 36.16 & 44.96 & 57.80 & 60.74 & 29.62 & 45.70 \\
Ours-MoE-Base-8k & 29.68 & 36.29 & 45.79 & 57.20 & 60.86 & 28.87 & 43.56 \\
Ours-MoE-SFT & \textbf{39.16} & \textbf{45.71} & \textbf{53.67} & \textbf{65.40} & \textbf{72.96} & \textbf{59.19} & \textbf{74.37} \\
\bottomrule
\end{tabular}
}
\caption{Performance comparison between our models and open-source models on different tasks.}
\label{tab:base-model-eval}
\end{table*}

\section{Experiments}

Due to the current lack of standardized Tibetan benchmarks for LLMs, we constructed four benchmarks to evaluate general Tibetan capabilities. Combined with two existing benchmarks for traditional Tibetan NLP tasks, this enables a comprehensive evaluation of model performance. Experimental results demonstrate that our model outperforms both open-source models of similar scale and Tibetan-tailored models across all evaluated tasks.

\subsection{General Tibetan Benchmark Construction}

Existing benchmarks~\cite{DBLP:journals/corr/abs-2502-20868, DBLP:conf/acl/LengJCHSPYXX25} for evaluating LLMs are predominantly designed for high-resource languages, with limited resources available for Tibetan. To facilitate systematic evaluation of Tibetan LLMs across a wide range of tasks, we constructed general-purpose Tibetan benchmarks based on the translation of several well-established English evaluation benchmarks, including HellaSwag~\cite{zellers2019hellaswag}, ARC\cite{clark2018think}, Xcopa~\cite{ponti2020xcopa}, and XStoryCloze~\cite{lin2021few}.


We leveraged the Claude-Sonnet-3.7 model, which demonstrates strong Tibetan generation capabilities, to translate the original datasets into Tibetan. To ensure the quality and reliability of the translated data, we performed a rigorous multi-stage review process covering semantic consistency, grammatical correctness, cultural appropriateness, and content safety. Detailed translation and quality review procedures are provided in Appendix~\ref{general-tibetan-benchmark-construction-details}. Ultimately, we constructed four benchmarks: HellaSwag-bo, ARC-bo, Xcopa-bo, and XStoryCloze-bo.

\subsection{Evaluation Benchmarks and Model}

We evaluated our model on a diverse set of Tibetan benchmarks, including our constructed tasks (HellaSwag-bo, ARC-bo, Xcopa-bo, XStoryCloze-bo) for reasoning and knowledge assessment, TibetanQA~\cite{sun2021construction} for reading comprehension, and FLORES-200~\cite{nllb2024scaling} for multilingual translation. Detailed benchmark descriptions are provided in Appendix~\ref{appendix:evaluation-benchmarks-details}.

%% file: section7-2.tex

\begin{table*}[ht]
	\centering
 \resizebox{\textwidth}{!}{%
	\begin{tabular}{lccccccccc}
\toprule
\multirow{2}{*}{\textbf{Model}}& \multicolumn{2}{c}{\textbf{Hellaswag-bo}}  & {\textbf{Arc-bo}} & {\textbf{Xcope-bo}} & \footnotesize{\textbf{Xstorycloze-bo}} & \multicolumn{2}{c}{\textbf{TibetanQA}}  &\multicolumn{2}{c}{\textbf{Flores-200}} \\
   & {(acc)} & {(acc\_norm)} & {(acc)} & {(acc)} & {(acc)} & {(EM)}& {(F1)} & {zh-bo}& {en-bo} \\\midrule
Ours-Base       & 30.30 & 36.20 & 44.32 & \textbf{59.80} &60.80 & \textbf{49.43} & \textbf{66.15} & 13.81 & 9.20  \\
Ours-Base-32k            & 30.09 & 36.58 & 44.62 & 58.60 & 60.60 & 38.42 & 55.52 & 20.86 & \textbf{13.56} \\
Ours-SFT & \textbf{30.64} & \textbf{36.71} & \textbf{48.39} & 57.80 & \textbf{61.86} & 46.15 & 63.15 & \textbf{21.56} & 11.42 \\ \midrule
Yak-Llama2-7B             & 24.67 & 26.40 & 25.64 & 53.00 & 48.37 & 0.12  & 0.24  & 11.81 & 11.10 \\
Tibetan-Alpaca-7B           & 25.95 & 28.23 & 27.94 & 51.20 & 49.97 & 2.47  & 4.81  & 4.87  & 5.69  \\
Tibetan-Llama2-7B           & 26.09 & 27.58 & 27.54 & 50.60 & 49.37 & 1.20  & 2.38  & 11.81 & 11.10      \\\bottomrule
	\end{tabular}
        }
	\caption{Performance of our models compared to other Tibetan models on different benchmarks.}
	\label{tab:tibetan}
\end{table*}

We conducted a comparative evaluation of three versions each of our dense models and MoE models. \textbf{Base} is trained through one stage of continuous pre-training. \textbf{Base-32/8k} undergo two stages. \textbf{SFT} have undergone further instruction fine-tuning. These models are evaluated against leading general-purpose open-source models, including the Qwen series~\cite{DBLP:journals/corr/abs-2412-15115, DBLP:journals/corr/abs-2505-09388}, LLaMA series~\cite{DBLP:journals/corr/abs-2407-21783}, and DeepSeek series~\cite{DBLP:journals/corr/abs-2501-12948}, on both standard reasoning and Tibetan-specific benchmarks.




\subsection{Main Results}

\paragraph{Dense Model.} As shown in Table~\ref{tab:base-model-eval}, on multiple-choice reasoning tasks such as Hellaswag-bo and ARC-bo, our models consistently outperform the baselines in accuracy scores. The advantage is even more pronounced on TibetanQA, where our models outperform general-purpose models by a large margin in both EM and F1 scores. These results underscore the effectiveness of our continued pretraining strategy, instruction tuning, and data construction pipeline in improving performance for low-resource languages like Tibetan.


\paragraph{MoE Model.} MoE-Base and MoE-Base-8k achieves performance comparable to the dense model across most benchmarks, indicating that MoE scaling does not compromise core Tibetan language understanding. After SFT, MoE-SFT exhibits substantial performance gains across all evaluated benchmarks. In particular, MoE-SFT significantly outperforms dense models on reasoning-intensive tasks such as XStoryCloze, and achieves a large margin improvement on TibetanQA, with the F1 score reaching 74.4. These results suggest that the expressive capacity introduced by MoE scaling is most effectively unlocked after alignment, enabling the model to better leverage both expanded parameters and Tibetan-specific data.

\subsection{Comparison against Existing Tibetan-Tailored Models}


We further compared our models with several instruction-tuned models specifically designed for Tibetan. As shown in Table~\ref{tab:tibetan}, our model consistently outperform existing Tibetan instruction-tuned models across all tasks. In particular, there are significant improvements in reading comprehension and question-answering tasks, while other models lag far behind in these tasks. Moreover, in both zh-bo and en-bo translation tasks, our models achieve the highest BLEU scores among all models, demonstrating strong multilingual alignment and effective cross-lingual generalization.


\subsection{Long Context Pre-training}
\label{long_context_pre-training_details}


\textbf{Data Distribution.} To fully account for the distributional characteristics of the training data in the first stage and further enhance the model's long context processing capabilities in Tibetan, we added long mixed-language texts (Chinese, English, and Tibetan) as training corpora in the second training stage.We sampled 0.56B tokens of Chinese data, 2.5B tokens of English data, and 1.5B tokens of Tibetan data (unused in the first stage) from long context sources in each respective language, resulting in a trilingual long context training set totaling 4.56B tokens.


\textbf{Training Parameters.} For training configurations, we maintained consistency with the Qwen2.5-7B-base model by setting the RoPE $\theta$ parameter to 1,000,000. The global batch size was set to 128 to ensure consistency in the number of tokens per batch with the first stage pre-training. Additionally, the learning rate was reduced from 5e-5 to 1e-5, while all other hyperparameters remained unchanged. This stage of training was conducted using DeepSpeed ZeRO-3 Offload strategy for a total of 1,000 steps.

\begin{figure*}[t]
    \centering

    \begin{subfigure}[b]{0.24\textwidth}
        \centering
        \includegraphics[width=\textwidth]{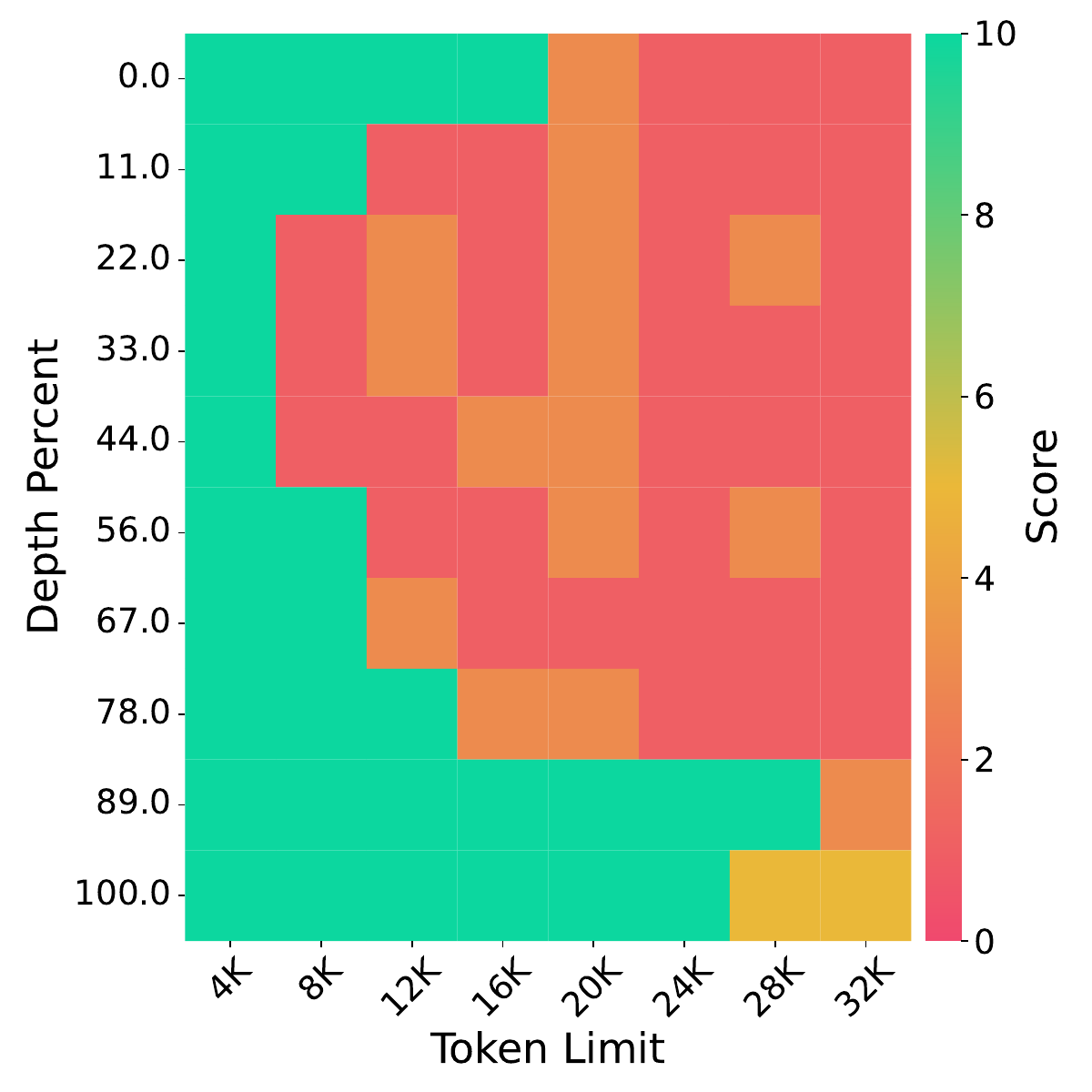}
        \caption{4K\_base/en}
        \label{fig:short_base_en}
    \end{subfigure}
    \begin{subfigure}[b]{0.24\textwidth}
        \centering
        \includegraphics[width=\textwidth]{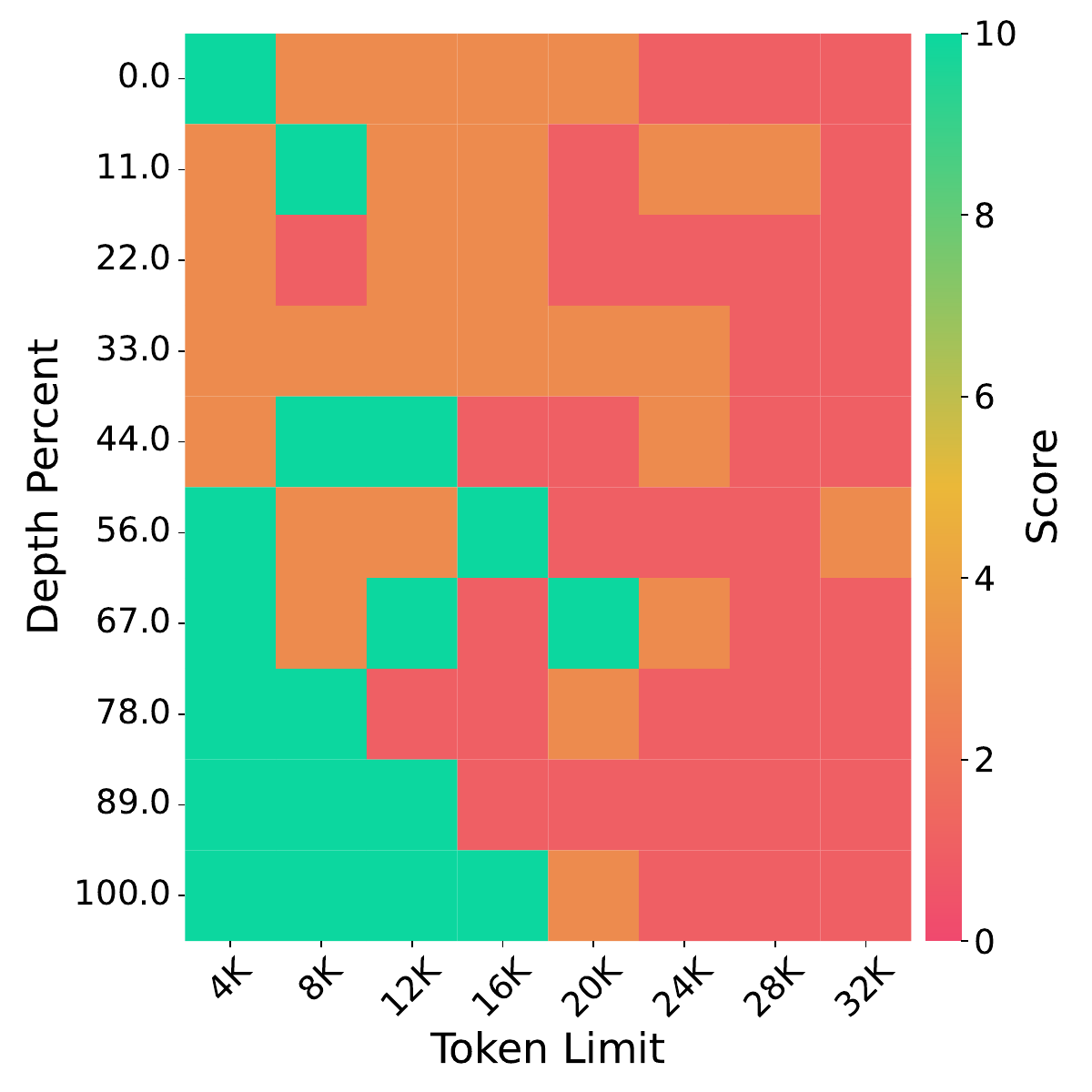}
        \caption{4K\_base/bo}
        \label{fig:short_base_ti}
    \end{subfigure}
    \begin{subfigure}[b]{0.24\textwidth}
        \centering
        \includegraphics[width=\textwidth]{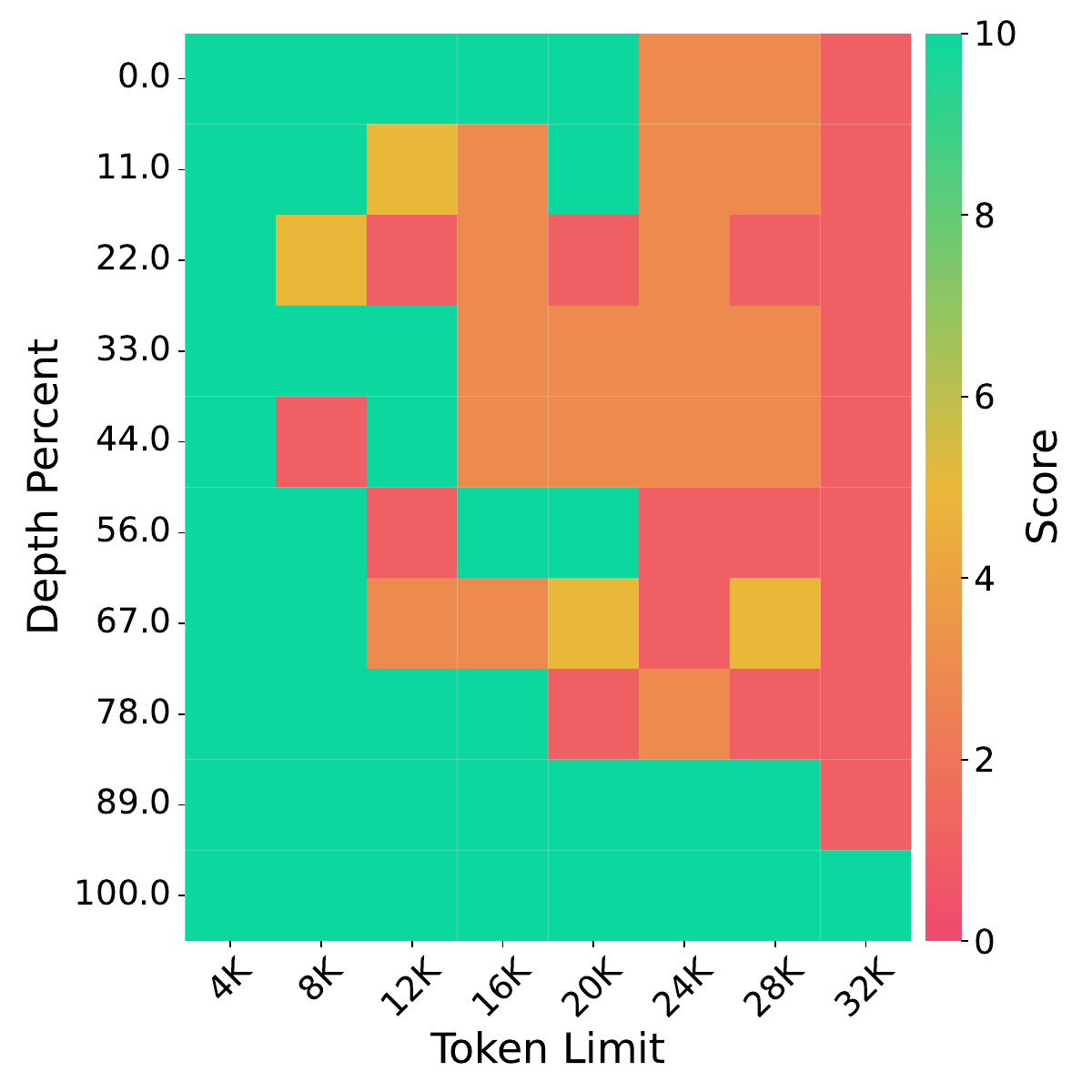}
        \caption{4K\_SFT/en}
        \label{fig:short_sft_en}
    \end{subfigure}
    \begin{subfigure}[b]{0.24\textwidth}
        \centering
        \includegraphics[width=\textwidth]{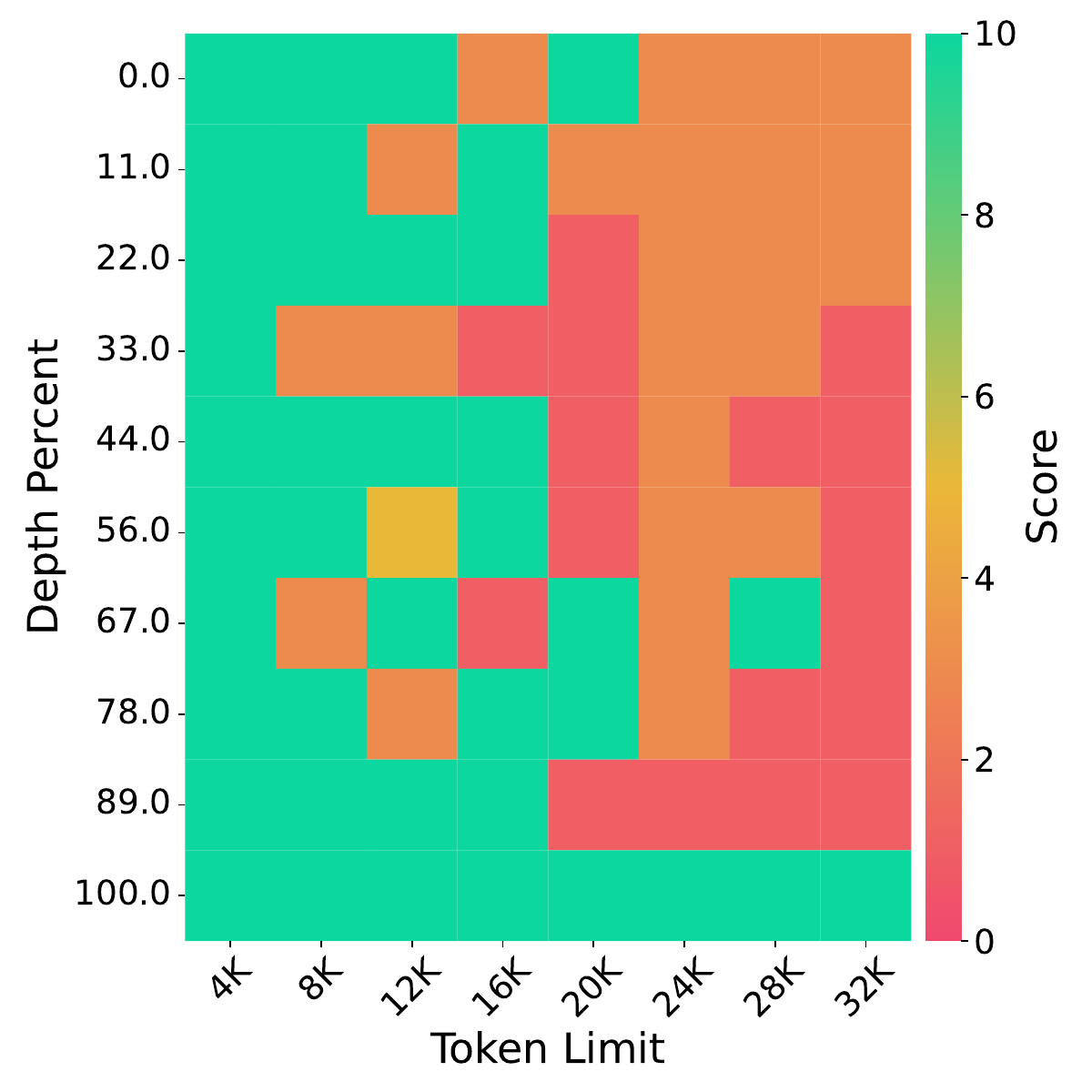}
        \caption{4K\_SFT/bo}
        \label{fig:short_sft_ti}
    \end{subfigure}

    \vspace{0.3cm} 

    \begin{subfigure}[b]{0.24\textwidth}
        \centering
        \includegraphics[width=\textwidth]{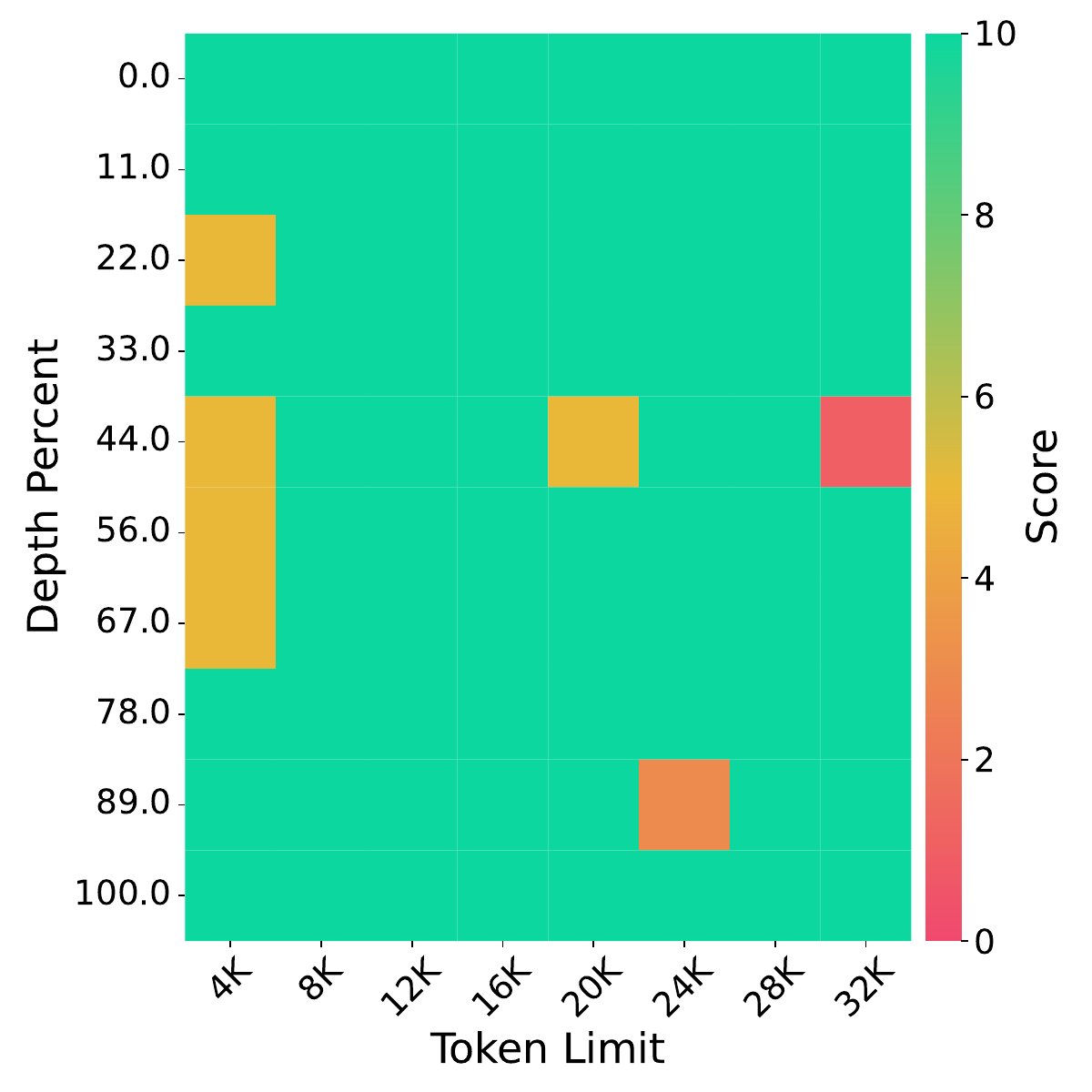}
        \caption{32K\_base/en}
        \label{fig:long_base_en}
    \end{subfigure}
    \begin{subfigure}[b]{0.24\textwidth}
        \centering
        \includegraphics[width=\textwidth]{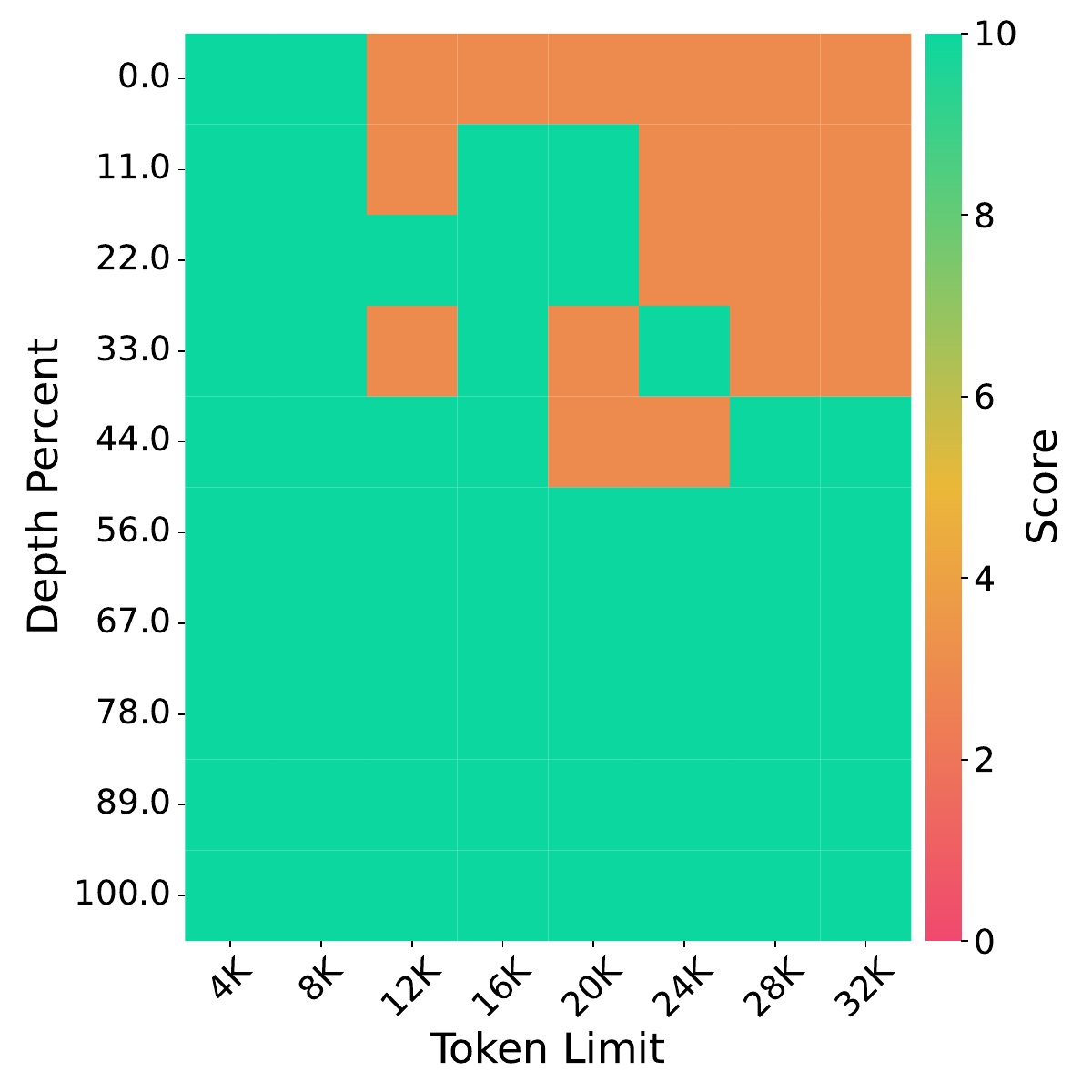}
        \caption{32K\_base/bo}
        \label{fig:long_base_ti}
    \end{subfigure}
    \begin{subfigure}[b]{0.24\textwidth}
        \centering
        \includegraphics[width=\textwidth]{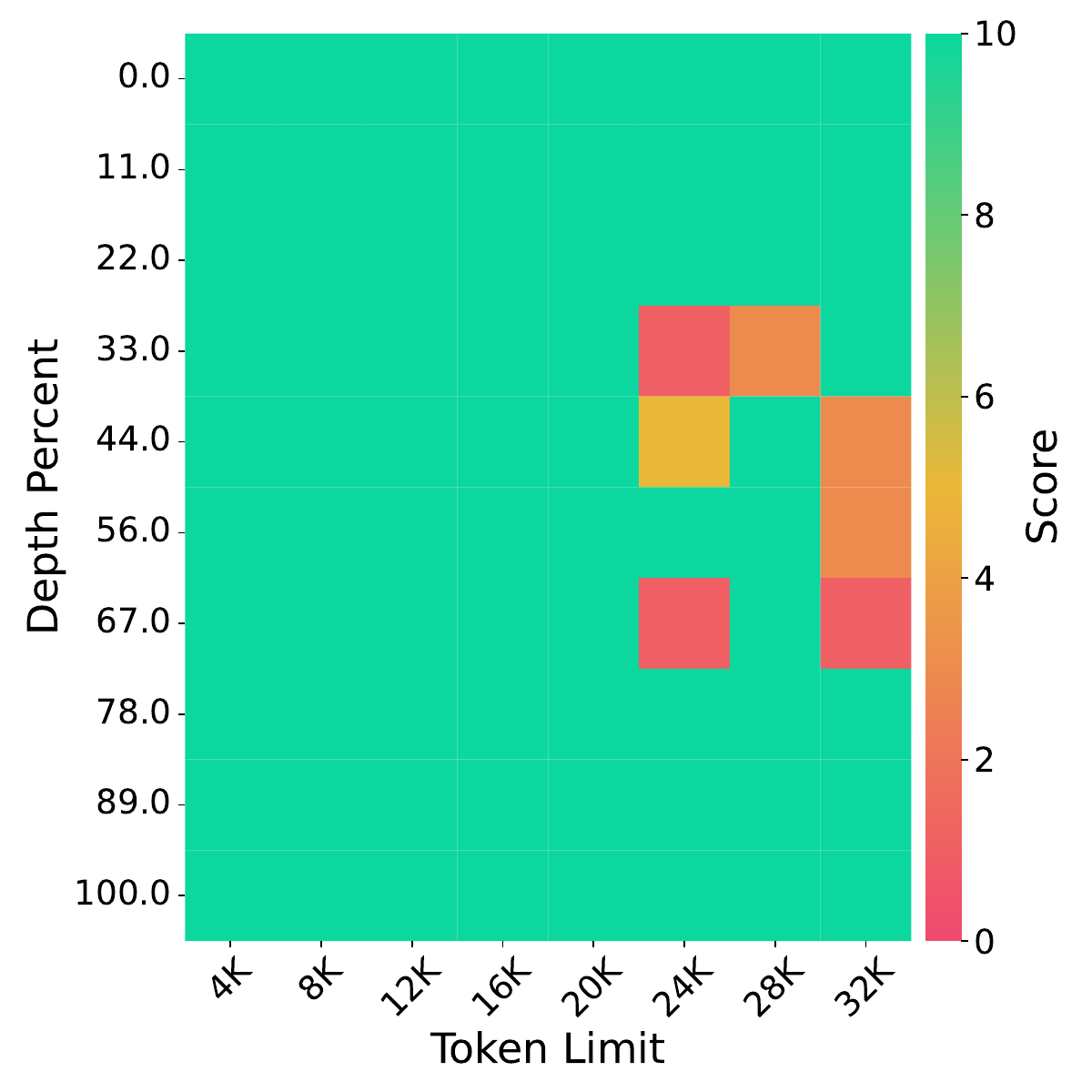}
        \caption{32K\_SFT/en}
        \label{fig:long_sft_en}
    \end{subfigure}
    \begin{subfigure}[b]{0.24\textwidth}
        \centering
        \includegraphics[width=\textwidth]{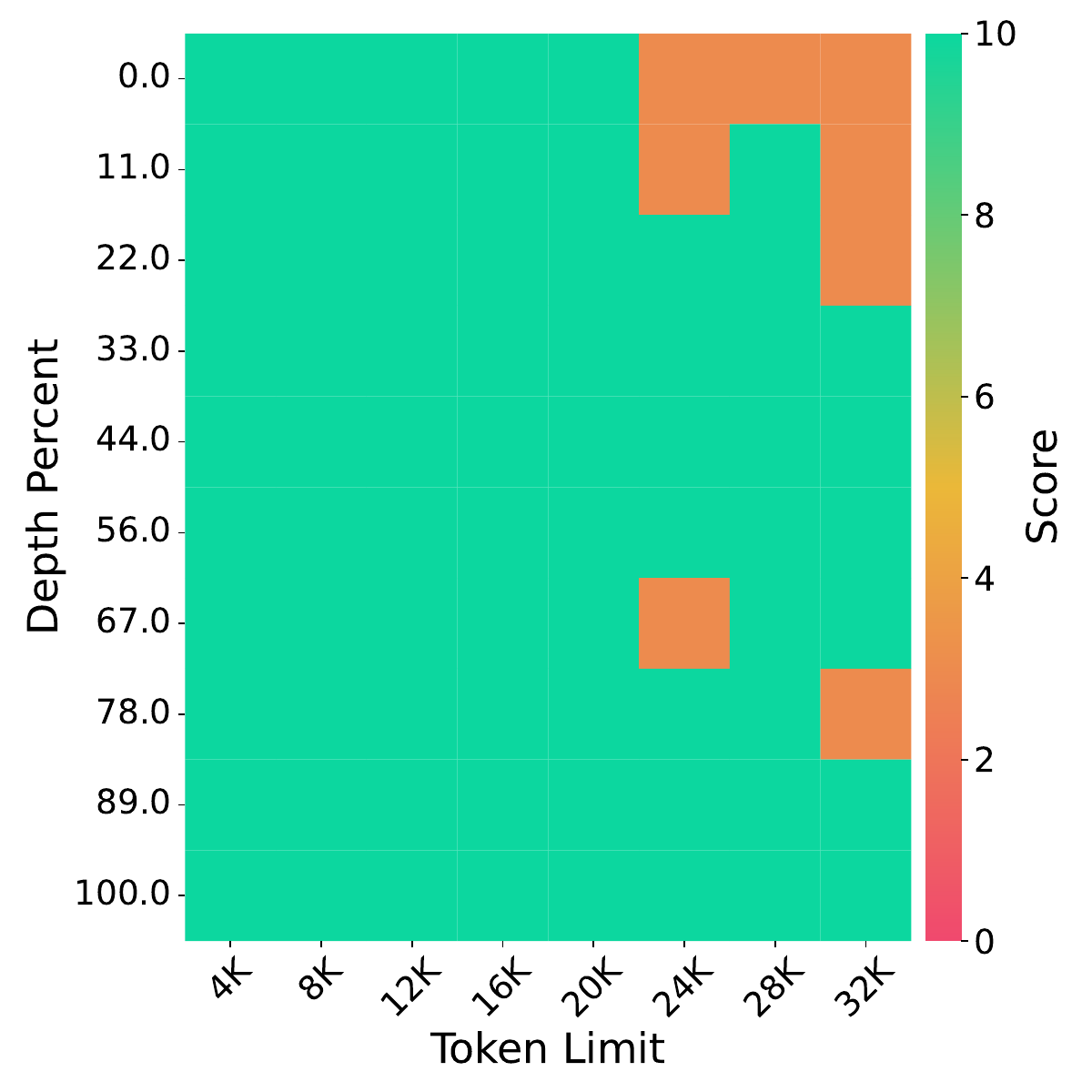}
        \caption{32K\_SFT/bo}
        \label{fig:long_sft_ti}
    \end{subfigure}

    \caption{Performance of the first stage 4K context window base and SFT models, as well as the second stage long context training 32K context window base and SFT models on the English (en) and Tibetan (bo) needle-in-a-haystack task.}
    \label{fig:niah}
\end{figure*}

\textbf{Experimental Results.} We evaluated the performance of four model variants on the needle-in-a-haystack task: the base and SFT models from the first stage, and the base and SFT models after long context training in the second stage. Evaluations were conducted using bilingual long context samples in English and Tibetan. As illustrated in Figure~\ref{fig:niah}, experiment results demonstrate that long context training successfully extend the model's context window to 32K tokens and significantly improve its multilingual capabilities.

\subsection{Tokenization Efficiency Analysis}
\label{sec:token}



In addition to accuracy-oriented evaluations, we analyze the efficiency of our tokenizer by measuring the tokenizer compression ratio, defined as the average number of characters per token. This metric reflects how efficiently a tokenizer represents text and directly affects training and inference cost, as well as the length of input text.



Specifically, our Tibetan model demonstrates exceptional performance, achieving a compression ratio of 3.9644. This far surpasses those of mainstream models developed in the industry, e.g., Qwen3-8B's 0.7315 and Llama3-8B's 0.4885, highlighting a significant advantage in compression capability. A high compression ratio indicates that when processing Tibetan texts of the same length, a model requires fewer tokens, which effectively reduces computational load and memory usage, thereby significantly accelerating the inference process. This feature not only enhances the efficiency of the model but also provides strong support for the application of Tibetan natural language processing tasks in real-time response scenarios and resource-constrained environments, underscoring the prominent value of our method in optimizing inference speed and reducing computational cost.

%% file: section8.tex
\section{Conclusion}



We have presented a practical and scalable framework for improving LLMs for Tibetan, a severely underrepresented yet culturally important low-resource language. By constructing the largest high-quality Tibetan pre-training corpus to date and applying a tailored data cleaning pipeline, we address key data scarcity and quality challenges. Through balanced multilingual continual pre-training, vocabulary expansion, and multilingual instruction tuning, our models achieve substantial improvements in Tibetan understanding and generation. We further demonstrate that MoE scaling, when initialized from a Tibetan-adapted dense model, offers an effective path to increasing model capacity without sacrificing efficiency. Finally, the introduction of new, carefully validated Tibetan benchmarks enables more systematic evaluation. Together, these contributions establish a strong foundation for Tibetan LLM research and provide transferable insights for extending LLMs to other low-resource languages.

%% file: section9.tex
\clearpage

\section*{Limitations}



Despite the strong empirical results, this work has several limitations. First, although the Tibetan corpus we construct is the largest to date, it still remains modest compared to corpora available for high-resource languages, which may limit the model’s coverage of rare domains, styles, and dialectal variations of Tibetan. Second, part of the training and evaluation data relies on machine translation and synthetic generation, which can introduce translation artifacts or subtle semantic biases despite careful filtering and human verification. Third, our evaluation benchmarks, while broader than existing resources, are primarily translated from English datasets and may not fully capture language-specific reasoning patterns or culturally grounded knowledge intrinsic to Tibetan. Finally, the Mixture-of-Experts analysis focuses mainly on performance outcomes; deeper investigation into expert specialization, routing behavior, and efficiency–quality trade-offs is left for future work. Addressing these limitations will require larger native Tibetan datasets, more culturally grounded benchmarks, and more detailed analyses of scalable architectures for low-resource languages.

\section*{Acknowledgments}

The present research was supported by the National Key Research and Development Program of China (Grant No. 2023YFE0116400), the State Key Laboratory of Tibetan Intelligence (Grant No. 2025-ZJ-J08), and the Postdoctoral Fellowship Program of CPSF (Grant No. GZC20251075). We would like to thank the anonymous reviewers for their insightful comments.

%% file: appendix.tex
\newpage




\section{Open-Source Data Collection}
\label{appendix:open-source-data-collection-details}


The open-source Tibetan corpus we collect primarily comes from two sources: (1) the Tibetan portions of multilingual open-source datasets, and (2) dedicated monolingual Tibetan datasets. These corpora cover a wide range of domains. However, we observe that a portion of the web-based data originates from different snapshots of Common Crawl, which may introduce redundancy. To address this, we apply deduplication to remove duplicate content across time. 

\section{Web Data Crawling}
\label{appendix:web-data-crawling-details}

We manually collect 45 Tibetan-language websites and extract textual data from them. Most of these sites are news-oriented, and our primary goal is to obtain up-to-date Tibetan text. During the web crawling process, we extract hyperlinks from each page and determine whether the linked page belongs to the same root domain. If so, and the page has not been previously visited, we recursively crawl the subpage.

\section{Quality Filtering}
\label{appendix:quality-filter-details}

\paragraph{Gopher Duplicate Filtering}

This method removes both documents with a large number of short, repeated segments and those with low duplication but excessively long content. We measure duplication at both the sentence and paragraph level by computing the proportion of duplicated content and the proportion of characters within duplicated segments.

\begin{itemize}
    \item \textbf{Character proportion in repeated n-grams}: For 2-gram, 3-gram, and 4-gram patterns, we compute the proportion of characters contained in the most frequent repeated n-gram relative to the total number of characters in the document, with filtering thresholds set at 0.20, 0.18, and 0.16, respectively. In contrast, for n-grams ranging from 5 to 10, we calculate the proportion of characters contained in all repeated n-grams relative to the total character count. The corresponding thresholds decrease gradually from 0.15 for 5-grams to 0.10 for 10-grams.
    \item \textbf{Character proportion in duplicated sentences}: Documents in which duplicated sentences contain more than 20\% of the total characters are filtered out.
    \item \textbf{Duplication ratio}: Documents are filtered out if more than 30\% of their sentences or 30\% of their paragraphs are exact duplicates.
\end{itemize}

\paragraph{Gopher Quality Filtering}

This stage filters out any documents that violate the quality heuristics defined in the Gopher data pipeline. Specifically, a document is removed if it meets any of the following conditions:

\begin{itemize}
    \item The document contains fewer than 50 words or more than 10,000 words.
    \item The document has an average word length less than 2 or greater than 10.
    \item The document is with a symbol-to-word ratio greater than 0.1.
    \item Fewer than 80\% of words in the document contain at least one alphabetic character.
    \item More than 90\% of sentences in the document begin with bullet points.
    \item More than 30\% of sentences in the document end with ellipses.
\end{itemize}

To better handle Tibetan text, we replace the default NLTK tokenizer\footnote{\href{https://github.com/nltk/nltk}{https://github.com/nltk/nltk}} with a custom tokenizer designed specifically for Tibetan. This allows for more accurate tokenization and filtering tailored to the unique linguistic features of Tibetan.

\paragraph{C4 Quality Filtering}

The following filtering rules are applied to ensure data cleanliness, in line with the C4 dataset preprocessing strategy:

\begin{itemize}
    \item Only lines containing at least three words are retained.
    \item Any page containing words from a predefined list of offensive, profane, obscene, or otherwise inappropriate terms is discarded.
    \item All lines containing the word ``Javascript” are removed.
    \item Any page containing the phrase ``lorem ipsum” is discarded.
    \item Any page containing curly braces is removed.
    \item Since some crawled pages originated from Wikipedia and included citation markers, all pages containing such citation patterns are excluded.
    \item To eliminate templated or boilerplate legal content, we remove any line that contains phrases such as “terms of use”, “privacy policy”, “cookie policy”, “uses cookies”, “use of cookie”, or “use cookie”.
\end{itemize}

\paragraph{FineWeb Quality Filtering}

Documents that violate the FineWeb quality heuristics are removed. Specifically, a document is filtered out if it meets any of the following conditions:

\begin{itemize}
    \item More than 67\% of its sentences have a length less than or equal to 30 characters.
    \item After removing empty lines, the proportion of repeated characters exceeds 1\% of the total character count.
    \item The ratio of newline characters to total word count exceeds 0.3.
\end{itemize}

\paragraph{Sensitive Content Filtering}

To eliminate sensitive content from the collected data, we manually curate a list of sensitive terms. Any document containing one or more terms from this list is filtered out.

\section{Continual Pre-training}
\label{continual-pre-training-details}


To improve training efficiency, we adopt a packing strategy~\cite{raffel2020exploring} to preprocess the data. This involves concatenating and truncating the data to ensure that each training sample is exactly 4096 tokens in length. Given the large volume of training data, we also apply a pre-tokenization strategy, where the tokenized and packed datasets are preprocessed and saved in advance to streamline the subsequent training process.

For the training framework, we use the LLaMA-Factory library~\cite{zheng2024llamafactory} in combination with DeepSpeed, enabling the ZeRO Stage 2 optimization strategy~\cite{rajbhandari2020zero}. Training is conducted using 32 A800 GPUs with bf16 precision. Regarding optimization, we use the standard AdamW optimizer~\cite{loshchilov2017decoupled} with a maximum gradient norm of 1.0. The learning rate is set to 5e-5, with a warmup ratio of 0.1, and we adopt a cosine learning rate scheduler.

\section{Post Training Details}
\label{post_training_details}

\paragraph{Training Data} We extensively collect open-source instruction datasets in Chinese, English, and Tibetan. Upon analysis, we find that most existing Tibetan instruction datasets are generated by translating open-source Chinese and English instruction data using Google Translate. While this approach is simple and convenient, it often results in low-quality Tibetan instructions due to the current limitations of Google's Tibetan translation capabilities. Additionally, such translations may introduce unintended biases. To address these issues, we collaborate with native Tibetan speakers to produce additional high-quality translation and summarization data. These samples are organized using instruction-style templates to further enhance the model’s instruction-following abilities in Tibetan, particularly in the areas of translation and summarization.

\paragraph{Training Setting} We use a packing strategy to process the instruction data and adopted the standard supervised fine-tuning loss computation method. Specifically, the loss is calculated only on assistant tokens, while system prompt and user tokens were excluded. The learning rate is set to 1e-5 while the maximum gradient norm 1.0. The model is trained for three epochs.

\section{General Tibetan Benchmark Construction}
\label{general-tibetan-benchmark-construction-details}

To construct general-purpose Tibetan benchmarks for evaluating LLMs, we translate several widely used English evaluation datasets (e.g., HellaSwag, ARC, XCOPA, and XStoryCloze) into Tibetan. These datasets span a range of reasoning, commonsense, and knowledge-intensive tasks, making them suitable for comprehensive capability assessment.

\paragraph{Translation} We use the Claude-Sonnet-3.7 model to perform English-to-Tibetan translation. This model is equipped with advanced capabilities for Tibetan generation, enabled by large-scale cross-lingual pre-training.

\paragraph{Quality Review} To ensure translation quality, we perform a structured manual review process guided by our internal Data Review Guidelines. The review procedure consists of three main stages:

\begin{itemize}
\item \textbf{Semantic and Terminological Accuracy:}
Reviewers compare the source and translated texts to identify semantic deviations, including incorrect interpretation of core concepts, logical inconsistencies, or mistranslation of technical terms.

\item \textbf{Content Compliance:}
All translated content is checked for potentially harmful material, such as violent or discriminatory language. Additionally, reviewers filter out content that may conflict with Tibetan cultural norms.

\item \textbf{Linguistic and Formatting Correctness:}
Reviewers examine grammatical structures (e.g., use of case particles, honorifics), fluency, and formatting consistency, including punctuation, line breaks, and symbol pairing.

\end{itemize}

Translations exhibiting conspicuous machine translation artifacts or unnatural expressions shall be rejected under the ``Low-Quality Machine Translation” category to ensure that our training data align with Tibetan linguistic conventions. During review, annotators must maintain objective assessment, comprehensively label rejection categories for multi-issue samples with specific problem descriptions, thereby constructing a high-quality Tibetan evaluation benchmarks.

\section{Evaluation Benchmarks}
\label{appendix:evaluation-benchmarks-details}

We used six key benchmarks, each designed to assess a specific capability of models within the Tibetan language setting:

\begin{itemize}
    \item HellaSwag-bo: Evaluates commonsense reasoning about event progression using multiple-choice questions. Accuracy is measured against adversarial distractors to test the model's ability to make logically coherent predictions across diverse scenarios.
    \item Xcopa-bo: Focuses on causal reasoning by requiring the model to choose the correct cause or effect of a given event.
    \item XstoryCloze-bo: Assesses zero-shot narrative understanding by requiring a model to choose the most plausible story ending. It evaluates both temporal consistency and semantic coherence in Tibetan.
    \item ARC-bo: Benchmarks factual knowledge by evaluating middle-school level science questions that require basic recall. Accuracy reflects the ability of a model to retrieve simple facts in Tibetan.
    \item TibetanQA: A large-scale machine reading comprehension dataset containing 20,000 QA pairs across 12 domains and four difficulty levels: word matching, synonym substitution, multi-sentence reasoning, and ambiguous questions. It tests fine-grained Tibetan text understanding.
    \item FLORES-200 (Tibetan subset): Used to evaluate translation quality among Chinese, English, and Tibetan. The dataset consists of professionally verified parallel sentences and measures character handling, grammatical correctness, and translation fidelity.
\end{itemize}